\PassOptionsToPackage{table}{xcolor}
\documentclass[conference]{IEEEtran}
\IEEEoverridecommandlockouts

\usepackage{cite}
\usepackage{cuted}
\usepackage{amsmath,amssymb,amsfonts}
\usepackage{graphicx}
\usepackage{textcomp}
\usepackage{url}
\usepackage{booktabs}

\usepackage{siunitx}
\usepackage{pgf}
\usepackage{hyperref}

\newcommand{\topfloatguard}{\vspace*{0.08in}}

\def\BibTeX{{\rm B\kern-.05em{\sc i\kern-.025em b}\kern-.08em
    T\kern-.1667em\lower.7ex\hbox{E}\kern-.125emX}}

\hypersetup{
  hidelinks,
  pdftitle={Preference-Conditioned Multi-Objective Reinforcement Learning for Runtime-Tunable Transit Signal Priority},
  pdfauthor={Philip-Roman Adam and Stefanie Schmidtner}
}

\begin{document}

\title{Preference-Conditioned Multi-Objective Reinforcement Learning for Runtime-Tunable Transit Signal Priority}

\author{
\IEEEauthorblockN{Philip-Roman Adam}
\IEEEauthorblockA{
\textit{Technische Hochschule Ingolstadt}\\
{Ingolstadt, Germany} \\
{philip-roman.adam@thi.de}
}
\and
\IEEEauthorblockN{Stefanie Schmidtner}
\IEEEauthorblockA{
\textit{Technische Hochschule Ingolstadt}\\
{Ingolstadt, Germany} \\
{stefanie.schmidtner@thi.de}
}
}

\maketitle

\begin{strip}
\centering
\small
\copyright~2026 IEEE. Personal use of this material is permitted.
Permission from IEEE must be obtained for all other uses, in any current
or future media, including reprinting/republishing this material for
advertising or promotional purposes, creating new collective works, for
resale or redistribution to servers or lists, or reuse of any copyrighted
component of this work in other works.
\end{strip}

\begin{abstract}
Transit signal priority (TSP) requires balancing competing objectives: reducing bus delay while limiting adverse impacts on non-bus traffic and avoiding extreme waits for a subset of vehicles. Existing reinforcement-learning (RL) approaches to TSP typically encode transit-aware features (e.g., occupancy and schedule deviation) but optimize a fixed reward or fixed scalarization, which limits operational flexibility when agency priorities change across time-of-day or disruption conditions. We present a preference-conditioned TSP controller, $\pi(a\mid s,w)$, that selects the next signal phase under minimum/maximum green and transition-feasibility constraints and can be tuned at runtime via a preference parameter $w$ to trade off bus-priority emphasis against overall traffic delay without retraining. We implement this on top of IntersectionZoo by introducing a constrained signal-control/TSP wrapper, and we extend scenario generation with bus-prevalence augmentation and timetable-based bus insertion to address sparse transit-priority events during training.
Experiments against fixed-time control, a rule-based TSP overlay, and fixed-weight PPO specialists show that a single learned conditioned policy spans a smooth empirical trade-off frontier across runtime preferences, outperforms fixed-time and rule-based baselines, and maintains constraint feasibility, while tail-delay diagnostics reveal that non-bus externalities remain limited for moderate preference settings but can increase substantially under high bus-priority weights. The source code of this work is available at \href{https://github.com/urbanAIthi/morl-tsp}{github.com/urbanAIthi/morl-tsp}.
\end{abstract}


\section{Introduction}
Transit signal priority (TSP) improves bus travel time and reliability by reallocating green time when buses approach signalized intersections, but it can impose externalities on non-bus traffic if applied aggressively. Because agency priorities vary across time of day and under disruptions, a single fixed ``priority setpoint'' is often operationally insufficient. In practice, classical TSP uses rule-based logic (e.g., green extension (GE), early green (EG), or phase insertion) triggered by bus detection and constrained by safety timing rules and controller capabilities~\cite{Smith2005TSPHandbook,NTCIP1211_2014}.

Recent reinforcement learning (RL) approaches to TSP incorporate transit-aware features such as bus occupancy and schedule deviation, but typically optimize a fixed reward scalarization, requiring retraining or retuning when the desired bus-priority vs.\ network-delay trade-off changes~\cite{LONG2022103814,Hu2023TwoWayTSP,LongChung2025CoopTSP,Zhou2025ConstrainedTSC}. A standard workaround is to train multiple fixed-weight specialists and select among them, but this yields only discrete operating points and increases training cost. In multi-objective RL (MORL), state-of-the-art weight-flexible control uses preference-conditioned value or policy networks that take a weight vector $w$ as input (e.g., UVFA/PCN-style conditioning), enabling continuous runtime tuning without retraining~\cite{Schaul2015UVFA,Abels2019ConditionedMO,Reymond2022PCN}. This flexible-objective perspective has not been systematically evaluated for TSP under shared operational constraints.

Empirical comparison in RL-based signal control remains uneven: studies vary substantially in simulator/sensing assumptions, intersection layouts, demand generation, and baseline implementations, and many do not compare within a common testbed, making state-of-the-art claims harder to interpret~\cite{Ault2021RESCO,CabrejasEgea2021TSC}. Benchmarking libraries were introduced explicitly to standardize scenarios and baseline implementations for fair comparisons (e.g., RESCO, LibSignal)~\cite{Ault2021RESCO,Mei2023LibSignal}.

We present a preference-conditioned TSP controller $\pi(a\mid s,w)$ for constrained phase selection on IntersectionZoo-derived (IZ) intersections \cite{Jayawardana2025IntersectionZoo}. We additionally introduce an IZ-to-TSP wrapper (shared feasibility layer,
bus-state extraction, and configurable bus prevalence) and evaluate runtime tunability via preference sweeps,
including tail-delay and fairness diagnostics to expose concentrated non-bus harms.

\subsection{Contributions}
We make three contributions:
\begin{itemize}
  \item \textbf{IZ--TSP benchmark suite (wrapper + baseline pack):} a standardized constrained phase-control TSP task on IZ-derived intersections with bus-state extraction, configurable bus prevalence/attributes, and a constraint-matched baseline pack (fixed-time + standard rule-based GE/EG overlay) to enable reproducible, comparable evaluation across methods.
  \item \textbf{Preference-conditioned TSP controller:} a single policy $\pi(a\mid s,w)$ enabling runtime tuning of bus vs.\ all-vehicle-delay trade-offs without retraining, evaluated across preference sweeps.
  \item \textbf{Trade-off and robustness protocol:} Pareto-style reporting across runtime preferences plus distribution-shift stress tests, with tail-delay/fairness diagnostics.
\end{itemize}

\section{Problem Formulation and Benchmark Design}\label{sec:problem}
We study TSP at a single signalized intersection with a fixed set of green phases $P$. At each decision step $t$, the controller observes $s_t$ and a runtime preference parameter $w$ that specifies the desired scalarization between bus delay and all-vehicle delay, and outputs a requested next green phase $a_t \in \{0, \ldots, |P|-1\}$. For request-based controllers (RuleTSP and learning-based methods), actions are subject to operational constraints (e.g., min/max green and transition feasibility) enforced by a shared feasibility layer (Section~\ref{sec:feas}) so that these methods are compared under identical constraints.

\subsection{Control Interface and Timing Constraints}\label{sec:timing}
All request-based controllers (RuleTSP and learning-based methods) interface with the same phase controller. Every $4\,\mathrm{s}$, the agent requests the next green phase, while transition phases are not selectable. FixedTime replays the scenario tlLogic program and is therefore not quantized to the $4\,\mathrm{s}$ decision grid. SUMO advances with a simulation step of $1\,\mathrm{s}$. The controller executes a switch only when the timing constraints are satisfied. In all experiments, we enforce a minimum green of $20\,\mathrm{s}$ and a maximum green of $50\,\mathrm{s}$. Control decisions are issued every $4\,\mathrm{s}$, so greens can only be extended in $4\,\mathrm{s}$ increments (the minimum extension). Otherwise, the controller keeps the current phase. Requests do not persist across decision steps. If a switch is not executed, the agent issues a new request at the next decision time. When a green-to-green switch is executed, the controller inserts a fixed yellow transition of $3\,\mathrm{s}$. No all-red stage is used, although the wrapper supports one.

\subsection{Observation}\label{sec:obs}
At each decision step $t$, the observation concatenates signal status, traffic conditions, and transit information for approaching buses. Signal features encode the active phase, time since the last phase change, time-to-min-green satisfaction, and the time since each phase was last active.

Traffic features are computed over vehicles within a fixed detection range of 250~m upstream of the stop line. Vehicles that have not yet cleared their previous signalized intersection are excluded. These features include lane-level density and queue estimates, and per-lane delay summaries. In particular, for each incoming lane $\ell$ we compute a tail-delay statistic $\mathrm{CVaR}^{\mathrm{lane}}_{0.1,\ell,t}$, defined as the mean per-vehicle delay among the worst-delay 10\% of vehicles currently observed on lane $\ell$.

Transit features use a fixed-size bus-slot encoding per incoming lane: for each incoming lane we allocate $K{=}2$ bus slots and encode each slot by the bus distance-to-stopline $d$ (keeping the closest buses and zero-filling empty slots). Distance-to-stopline is the lane/path distance to the controlled TLS, clipped to $[0,250~\mathrm{m}]$ and normalized. In the reported experiments, occupancy and schedule-deviation metadata are available in the wrapper for future transit-aware controllers, but are not included in the observation vector in order to isolate preference-conditioned delay control. This fixed-size slot encoding handles variable numbers of approaching buses and follows prior RL--TSP practice (e.g., selecting up to top-$K$ buses per movement and zero-padding the remainder)~\cite{LONG2022103814}.

\subsection{Objectives, Reward Vector, and Preference Conditioning}\label{sec:reward}
We evaluate controllers using episode-level delay metrics that quantify bus delay and non-bus delay as the main externality measure. All delay quantities are computed within the controlled detection region, defined as a fixed upstream segment of length 250\,m for both buses and non-bus vehicles. Delay is measured as excess travel time relative to free flow, which is computed from the distance traveled inside the controlled detection region and the corresponding edge speed limits.

Let $B$ denote buses and let $V$ denote all vehicles. Let $\mathrm{delay}_k$ be the accumulated delay for vehicle $k$ in an episode. The episode-level evaluation metrics reported in tables and figures are mean per-vehicle delays:

\[
J_b = \frac{1}{|B|}\sum_{k\in B}\mathrm{delay}_k
\tag{1}
\]
\[
J_{nb} = \frac{1}{|V\setminus B|}\sum_{k\in V\setminus B}\mathrm{delay}_k .
\tag{2}
\]
We also compute the all-vehicle mean delay $J_{\mathrm{all}}$ analogously over $V$.

To quantify disproportionate impacts on non-transit users, we additionally report a non-bus tail-delay diagnostic based on $\mathrm{CVaR}_{\alpha}$ computed over the episode-delay distribution of vehicles in $V\setminus B$.

We learn a single conditioned policy
\begin{equation}
\pi(a \mid s,w)
\tag{3}
\end{equation}
In our experiments we use a single bus priority weight $w_{\mathrm{bus}}\in[0,1]$ with a complementary weight on the all-vehicle delay proxy. We vary $w$ at evaluation time to trace the empirical trade-off without retraining~\cite{Schaul2015UVFA,Abels2019ConditionedMO,Reymond2022PCN}.

For training, we use a two-dimensional per-step reward vector with bus-delay and all-vehicle-delay components. The all-vehicle component is additionally shaped with a non-bus tail-delay penalty based on $\mathrm{CVaR}_\alpha$ to discourage extreme waits. Concretely, $d^{bus}_t$ and $d^{all}_t$ are computed as sums of instantaneous in-region vehicle delays over, respectively, buses and all vehicles currently in the detection region at decision epoch $t$. Let $D^{\mathrm{nb}}_t$ denote the corresponding set of instantaneous non-bus vehicle delays, and define

\[
\mathrm{CVaR}^{\mathrm{nb}}_{\alpha,t} \triangleq \mathrm{CVaR}_{\alpha}\!\left(D^{\mathrm{nb}}_{t}\right),
\]
the empirical CVaR of the instantaneous non-bus delay distribution.

The per-step reward vector is
\begin{equation}
\mathbf{r}^{\text{vec}}_t =
\left[
\begin{array}{l}
-d^{\text{bus}}_t \\
-\left(d^{\text{all}}_t + \lambda_{\text{cvar}}\max\!\big(0,\mathrm{CVaR}^{\mathrm{nb}}_{\alpha,t}-T_{\text{cvar}}\big)\right)
\end{array}
\right].
\tag{4}
\end{equation}
We use $\alpha{=}0.1$, $T_{\text{cvar}}{=}120\,\mathrm{s}$, and $\lambda_{\text{cvar}}{=}0.25$ in all experiments.

To stabilize learning and improve coverage of the trade-off surface, we use a simple preference curriculum. We sample $w_{\mathrm{bus}} \sim \mathcal{U}(0.1,0.9)$ for the first 150k steps, then $w_{\mathrm{bus}} \sim \mathcal{U}(0,1)$ thereafter.

\subsection{Scenario Regimes and Wrappers}\label{sec:regimes}
Our implementation builds on SUMO-RL as the environment scaffold for SUMO-based signal control, and extends it with IZ-derived scenario loading, bus metadata injection, and the TSP-specific observation/reward interfaces used in this paper~\cite{sumorl,Jayawardana2025IntersectionZoo,SUMO}. In default IZ scenarios, buses are sparse relative to private vehicles ($0.35\%$), so priority events are rare and conflicting requests are uncommon. This reduces learning signal for priority arbitration. We therefore augment training scenarios to increase the frequency of bus priority events while preserving the underlying intersection geometry and signal definitions.

We provide two scenario wrappers. The first increases bus prevalence by reweighting the vehicle-type mixture. The second injects timetable-based bus services and attaches per-bus metadata, including route identity, the number of concurrently active routes, passenger occupancy, and schedule deviation. All wrapper parameters and sampling distributions are specified in the released configuration files (supplementary material) to ensure exact reproducibility.

For \emph{in-distribution} evaluation, we train and test under the same timetable-augmented regime. Specifically, we sample $4$ concurrent bus routes per episode and augment the base demand with these services. For each sampled route, headways are drawn uniformly from $[300,900]$\,s and departure times are perturbed by a uniform offset in $[-120,600]$\,s. Route durations are sampled uniformly from $[10{,}800,36{,}000]$\,s, and route definitions may be reused across episodes. Consequently, both the active-route set and its timetable realization vary across episodes. To keep total demand approximately constant, we remove any buses present in the underlying scenario and rescale the remaining traffic so that the overall vehicle count is preserved.

We additionally conduct \emph{distribution shift} experiments (Section~\ref{sec:shift}) using the default IZ scenario generation (i.e., without timetable augmentation), which yields substantially lower bus prevalence. This tests whether runtime tunability transfers to less transit-saturated settings. The protocol is analogous to robustness evaluation via domain randomization in simulation-based RL~\cite{DomainRandomization2017,MullerSabatelli2023TrafficDR,GeneraLight2020}.

\subsection{Feasibility Handling and Guardrails}\label{sec:feas}
All request-based controllers share a feasibility layer that implements the execution rules and timing constraints in Section~\ref{sec:timing}. If an agent proposes an infeasible phase change, the controller executes a no-op (keeps the current green) and the request does not carry over to the next decision step. For PPO, we mask infeasible actions from the policy's action set. For MORL, infeasible actions are not masked; instead, we apply a fixed infeasibility penalty \(P_{\mathrm{inf}}=120.0\) to the all-vehicle reward component to discourage repeated invalid requests.

\section{Controllers}\label{sec:controllers}
We evaluate two non-learning baselines (FixedTime, RuleTSP) and two learning-based controllers (PPO specialists, preference-conditioned MORL). All controllers share the same green phase set. Learning-based controllers (and RuleTSP) interact with the signal through the non-persistent phase-request interface and feasibility layer described in Sections~\ref{sec:timing}--\ref{sec:feas}. In contrast, FixedTime replays the pre-timed SUMO traffic light program provided by the underlying IZ scenario.

\subsection{Fixed-Time}\label{sec:ft}
The FixedTime baseline replays the scenario-provided SUMO traffic light program, which corresponds to an offline-optimized fixed-time plan obtained by exhaustively searching over candidate fixed-time timing plans and selecting the best plan for each intersection, following Thunig et al.~\cite{Thunig2019FixedTimeOpt,Jayawardana2025IntersectionZoo}. For comparability of intergreen handling, our learning-based controllers use the same transition convention (fixed yellow insertion and no all-red stage) implemented in the shared feasibility layer (Sections~\ref{sec:timing}--\ref{sec:feas}).

\begin{table}
\topfloatguard
\caption{Rule-Based TSP Baseline Parameters}
\label{tab:rulebased}
\centering
\begin{tabular}{l r}
\toprule
Parameter & Value \\
\midrule
Detection range & 250 m \\
Max ETA & 25 s \\
Min speed & 1.0 m/s \\
Min delay & 0.0 s \\
Early green enabled & True \\
Green extension enabled & True \\
Early-green ETA threshold & 18 s \\
Forced switch remaining & 2.0 s \\
Green extension target & 8.0 s \\
\bottomrule
\end{tabular}
\end{table}

\begin{figure*}
  \centering
  \includegraphics[width=\textwidth]{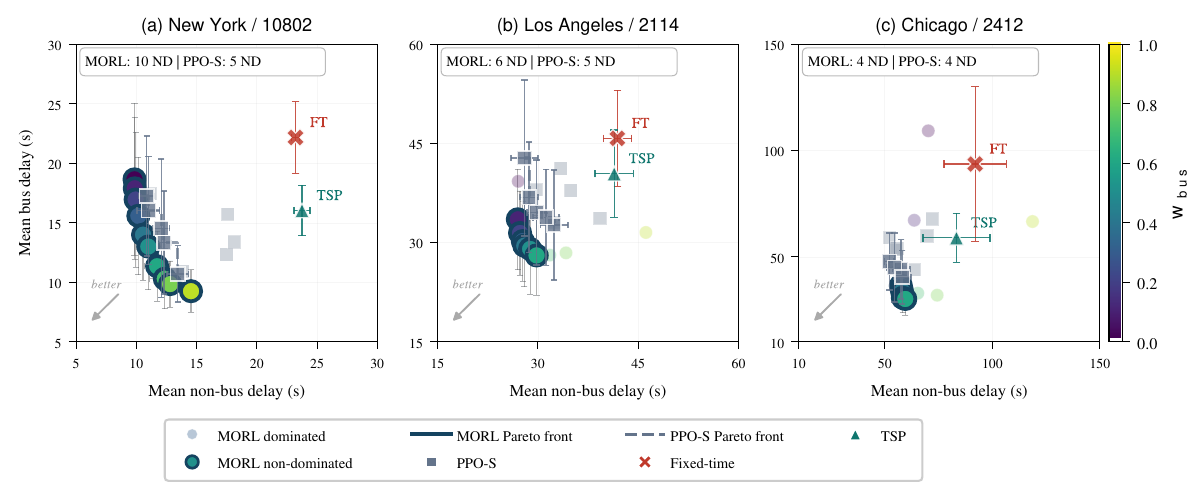}
  \caption{Pareto trade-off summaries across three case-study intersections (NYC10802, LA2114, CHI2412). Points show MORL mean outcomes across the 11-point preference sweep. Non-dominated points and the empirical Pareto front are highlighted, with fixed-weight PPO, fixed-time and rule-based TSP baselines for reference. Error bars denote mean $\pm$ one standard deviation across the 10 evaluation seeds. The ND counts shown inside each panel are computed on the seed-averaged points used for visualization; \autoref{tab:case_studies_wide} reports the mean~$\pm$~std of per-seed ND counts.}
  \label{fig:pareto_tradeoff}
\end{figure*}

\subsection{Rule-Based TSP Overlay (Fixed-Time + TSP)}\label{sec:ruletsp}
We implement a rule-based TSP overlay on fixed-time that applies GE when the current phase serves a qualifying bus or requests EG when the bus is within an ETA threshold. Eligibility uses a detection range plus delay/ETA/phase-compatibility checks. When multiple buses qualify, we select the smallest ETA. All requests are executed under the shared feasibility layer (Section~\ref{sec:feas}) with parameters fixed across experiments (Table~\ref{tab:rulebased}); these were selected based on standard GE/EG-style TSP settings from the literature and a small pilot sweep on CHI758, and we then held them fixed across all reported experiments to avoid per-intersection overtuning.

\subsection{Fixed-Weight PPO Specialists}\label{sec:ppo}
We train specialist RL controllers under fixed preference weights to optimize the scalarized reward. These state-of-the-art setpoint baselines are trained independently for $w_{\mathrm{bus}} \in \{0.0, 0.1, \ldots, 1.0\}$ and quantify the specialization versus generalization trade-off.

We use MaskablePPO to stabilize training by handling invalid action selection through an action mask derived from the shared feasibility layer (Section~\ref{sec:feas}), preventing selection of infeasible phase-change requests \cite{stable-baselines3,PPO}. As a result, infeasible-request penalties are not applied to PPO. We use the observation encoding in Section~\ref{sec:obs}.
The agents use an MLP policy with two hidden layers of width 256. Training uses rollouts of 1024 steps, minibatches of 128, learning rate $3\times 10^{-4}$, discount factor $\gamma=0.99$, and GAE parameter $\lambda=0.99$. We use a clipping range of 0.35, entropy coefficient 0.001, value-function coefficient 0.25, and 10 epochs per update. Each specialist is trained for 500k environment decision steps in total, aggregated across four vectorized environments.

\subsection{Preference-Conditioned MORL Controller}\label{sec:morl}
Our primary controller is trained with Envelope Q-Learning \cite{Yang2019Envelope} as implemented in \texttt{morl-baselines} \cite{MORLBaselines2023}. The method learns a single preference-conditioned action-value function $Q(s,a,w)$ and selects actions by maximizing $w^\top Q(s,a,w)$. This yields a single policy that can be tuned at runtime by changing $w$ without retraining.

The Q network is an MLP with 4 hidden layers of width 256 and ReLU activations. We use the same observation encoding as in Section~\ref{sec:obs}. We do not mask infeasible actions, so infeasible requests incur the penalty described in Section~\ref{sec:feas}. As hyperparameters we use a replay buffer of 250k, batch size 128, learning rate $5\times10^{-5}$, $\gamma=0.975$, Polyak $\tau=0.05$, and a target-network update every 500 steps. We perform 3 gradient updates per environment step after a 4k-step warm start and clip gradients to a max norm of 1.0. Exploration follows an $\epsilon$-greedy schedule with $\epsilon$ decaying from 0.1 to 0.02 over 150k steps. Preference conditioning samples $16$ weights per update. We do not use prioritized replay. We apply a homotopy schedule with $\lambda$ increasing from 0.0 to 0.3 over 200k steps and train for 500k environment steps per intersection.

\subsection{Specialists and Conditioned Policy Comparison}\label{sec:specvsmorl}
We train specialist PPO controllers at fixed weights and compare them to the single preference-conditioned controller across a sweep of runtime preferences. This directly measures the specialization versus generalization trade-off.

Specialists require training 11 separate PPO policies (one per weight), whereas the MORL approach trains a single preference-conditioned policy. We therefore report performance on a per-policy basis and do not claim a matched \emph{total} training budget between the two approaches. Instead, we emphasize the practical benefit of runtime tunability without retraining.

\begin{table*}
\topfloatguard
\centering
\scriptsize
\setlength{\tabcolsep}{2.4pt}
\renewcommand{\arraystretch}{1.15}
\caption{Case-study intersections (evaluation protocol in Section~\ref{sec:exp}). Frontier quality is summarized by hypervolume (HV\%) and the number of non-dominated points (\#ND) over the 11-point preference sweep. FT-dom and TSP-dom report the mean$\pm$std number of preferences (out of 11) where the policy dominates the FixedTime or RuleTSP baseline. We also report operating-point mean delays (s) at $w_{\mathrm{bus}}\in\{0.3,0.5,0.7\}$. $J^{\mathrm{tail}}_{\mathrm{nb}}$ denotes the non-bus episode tail-delay metric (defined in Section~\ref{sec:exp})}
\label{tab:case_studies_wide}
\resizebox{\textwidth}{!}{%
\begin{tabular}{llrrrr|rrr|rrr|rrr}
\toprule
& & \multicolumn{4}{c|}{Frontier quality} & \multicolumn{3}{c|}{$w_{\mathrm{bus}}=0.3$} & \multicolumn{3}{c|}{$w_{\mathrm{bus}}=0.5$} & \multicolumn{3}{c}{$w_{\mathrm{bus}}=0.7$} \\
Tag & Method & HV\% & \#ND & FT-dom & TSP-dom & $J_b$ & $J_{\mathrm{nb}}$ & $J^{\mathrm{tail}}_{\mathrm{nb}}$ & $J_b$ & $J_{\mathrm{nb}}$ & $J^{\mathrm{tail}}_{\mathrm{nb}}$ & $J_b$ & $J_{\mathrm{nb}}$ & $J^{\mathrm{tail}}_{\mathrm{nb}}$ \\
\midrule
chi2412 & MORL & 83.7$\pm$1.3 & 2.9$\pm$1.4 & 8.3$\pm$1.3 & 6.9$\pm$1.0 & 36.4$\pm$4.8 & 57.4$\pm$2.2 & 165.0$\pm$3.0 & 30.7$\pm$7.1 & 58.3$\pm$2.4 & 173.7$\pm$4.4 & 32.9$\pm$7.9 & 65.3$\pm$3.5 & 198.0$\pm$9.0 \\
 & PPO-S & 83.1$\pm$2.2 & 3.2$\pm$0.8 & 10.0$\pm$2.2 & 7.0$\pm$4.3 & 47.8$\pm$13.5 & 52.1$\pm$2.3 & 153.7$\pm$5.8 & 44.9$\pm$16.3 & 54.4$\pm$3.7 & 152.6$\pm$6.2 & 43.9$\pm$10.1 & 63.6$\pm$5.2 & 168.9$\pm$9.6 \\
 & FixedTime & -- & -- & -- & -- & 93.7$\pm$36.3 & 92.0$\pm$14.5 & 295.0$\pm$83.5 & 93.7$\pm$36.3 & 92.0$\pm$14.5 & 295.0$\pm$83.5 & 93.7$\pm$36.3 & 92.0$\pm$14.5 & 295.0$\pm$83.5 \\
 & RuleTSP & -- & -- & -- & -- & 58.9$\pm$11.6 & 83.3$\pm$15.6 & 253.3$\pm$74.1 & 58.9$\pm$11.6 & 83.3$\pm$15.6 & 253.3$\pm$74.1 & 58.9$\pm$11.6 & 83.3$\pm$15.6 & 253.3$\pm$74.1 \\
\addlinespace
la2114 & MORL & 89.6$\pm$1.2 & 4.6$\pm$1.8 & 8.9$\pm$0.6 & 8.5$\pm$0.5 & 30.3$\pm$6.1 & 27.7$\pm$0.7 & 88.6$\pm$2.0 & 29.1$\pm$6.9 & 28.8$\pm$1.3 & 90.7$\pm$2.9 & 28.2$\pm$5.9 & 31.8$\pm$1.5 & 98.5$\pm$3.7 \\
 & PPO-S & 88.6$\pm$1.5 & 3.2$\pm$1.4 & 8.8$\pm$1.9 & 7.8$\pm$2.2 & 41.2$\pm$10.1 & 33.4$\pm$1.1 & 96.6$\pm$3.8 & 32.7$\pm$8.3 & 32.4$\pm$2.0 & 94.8$\pm$6.7 & 33.1$\pm$8.2 & 32.4$\pm$1.8 & 99.0$\pm$4.9 \\
 & FixedTime & -- & -- & -- & -- & 45.8$\pm$7.3 & 41.9$\pm$2.1 & 111.1$\pm$5.3 & 45.8$\pm$7.3 & 41.9$\pm$2.1 & 111.1$\pm$5.3 & 45.8$\pm$7.3 & 41.9$\pm$2.1 & 111.1$\pm$5.3 \\
 & RuleTSP & -- & -- & -- & -- & 40.4$\pm$6.6 & 41.4$\pm$2.9 & 110.4$\pm$8.9 & 40.4$\pm$6.6 & 41.4$\pm$2.9 & 110.4$\pm$8.9 & 40.4$\pm$6.6 & 41.4$\pm$2.9 & 110.4$\pm$8.9 \\
\addlinespace
nyc10802 & MORL & 96.1$\pm$0.6 & 7.3$\pm$1.3 & 9.3$\pm$1.3 & 7.4$\pm$3.0 & 16.6$\pm$5.5 & 10.1$\pm$0.4 & 57.9$\pm$1.2 & 14.0$\pm$4.0 & 10.9$\pm$1.0 & 60.1$\pm$1.3 & 11.3$\pm$3.1 & 12.3$\pm$1.3 & 61.5$\pm$2.3 \\
 & PPO-S & 95.7$\pm$0.8 & 5.3$\pm$1.3 & 9.8$\pm$0.6 & 6.9$\pm$2.9 & 17.0$\pm$4.7 & 11.0$\pm$0.9 & 59.6$\pm$2.6 & 13.4$\pm$2.3 & 17.4$\pm$0.6 & 65.5$\pm$1.2 & 14.4$\pm$5.0 & 12.2$\pm$0.6 & 62.9$\pm$2.2 \\
 & FixedTime & -- & -- & -- & -- & 23.2$\pm$4.6 & 23.2$\pm$0.3 & 68.9$\pm$1.4 & 23.2$\pm$4.6 & 23.2$\pm$0.3 & 68.9$\pm$1.4 & 23.2$\pm$4.6 & 23.2$\pm$0.3 & 68.9$\pm$1.4 \\
 & RuleTSP & -- & -- & -- & -- & 17.0$\pm$3.6 & 23.7$\pm$0.7 & 70.4$\pm$2.5 & 17.0$\pm$3.6 & 23.7$\pm$0.7 & 70.4$\pm$2.5 & 17.0$\pm$3.6 & 23.7$\pm$0.7 & 70.4$\pm$2.5 \\
\bottomrule
\end{tabular}}
\end{table*}

\section{Experimental Protocol}\label{sec:exp}
\subsection{Intersections and Selection Strategy}
We randomly selected 13 IZ-derived intersections with more than seven lanes. We use three intersections for detailed case studies and head-to-head comparison against the fixed-weight PPO specialists (NYC10802, LA2114, CHI2412). The remaining intersections are used to evaluate reproducibility and robustness across diverse layouts, namely ATL66, BOS666, CHI1027, CHI758, DAL1082, LA477, NYC1078, SEA543, SF838, and SLC47.

Bus-prevalence regimes and distribution-shift definitions are provided in Section~\ref{sec:regimes}.

All reported results use SUMO v1.25.0. IZ scenarios are generated using IntersectionZoo at revision \texttt{912d102}, which is included in the released repository as a pinned Git submodule.

\subsection{Training and Evaluation Protocol}
We train the preference-conditioned Envelope controller for 500k environment decision steps using a single environment. For each fixed preference weight, we train one PPO specialist for 500k total environment decision steps aggregated across four vectorized environments. Hyperparameters and the preference curriculum are tuned on CHI758 and then held fixed for all remaining intersections.

We train under the augmented timetable-insertion regime described in Section~\ref{sec:regimes}, where the active-route set and timetables are resampled each episode to induce continually varying, multi-route priority demand. At evaluation, for each method and each preference setpoint (or specialist), we run one 3-hour episode for each of 10 seeds (3423--3432) with randomized start times; each seed resamples both the active-route set and the timetable realization. This stresses arbitration under changing priority demand and tests generalization to previously unseen route combinations.

\subsection{Evaluation Metrics}

Training uses the per-step reward components defined in Section~\ref{sec:reward}, scalarized by the relevant preference weights where applicable, and we report episode-level mean per-vehicle bus and non-bus delays for externality analysis. Unless otherwise stated, Pareto-front summaries (HV, \#ND, dominance/coverage) are computed in the reported metric space \((J_b, J_{nb})\) over an 11-point preference sweep, aggregating uncertainty across 10 random seeds.

To characterize concentrated non-bus impacts, we additionally report an episode-level tail-delay metric
\[
J^{\mathrm{tail}}_{\mathrm{nb}} := \mathrm{CVaR}_{0.1}\!\left(\{\mathrm{delay}_k\}_{k \in V\setminus B}\right),
\]
i.e., the mean episode delay among the worst-delay 10\% of non-bus vehicles in the episode.

We evaluate 11 setpoints $w_{\mathrm{bus}} \in \{0.0,0.1,\ldots,1.0\}$. Hypervolume uses reference point $(500,500)$ in \((J_b,J_{nb})\) and is reported as HV\% normalized by $500{\times}500$.

\begin{figure*}
  \centering
  \includegraphics[width=\textwidth]{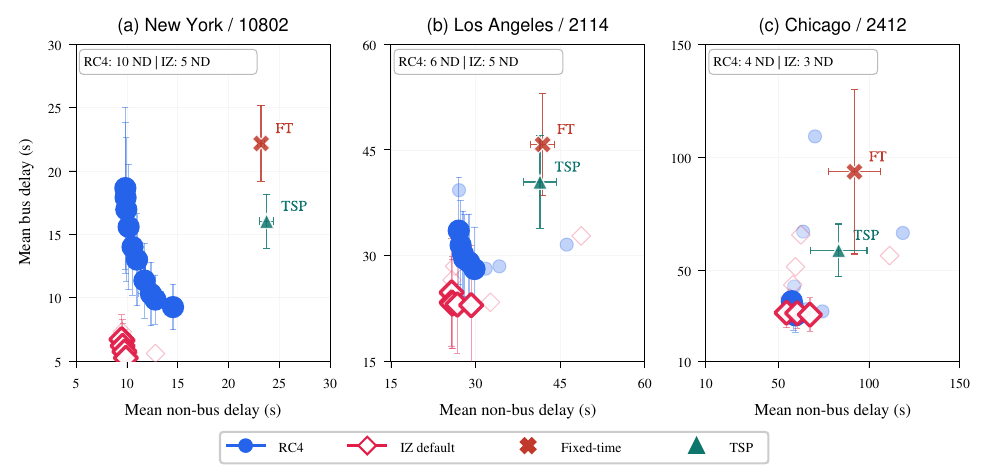}
  \caption{Distribution-shift evaluation (see Section~\ref{sec:regimes}) across NYC10802, LA2114, and CHI2412. RC4 denotes the in-distribution timetable-augmented regime with four concurrent bus routes; default-IZ denotes the lower-bus-prevalence regime without timetable augmentation. The panels compare MORL Pareto fronts under both regimes, with fixed-time and rule-based TSP baselines shown for reference. Error bars denote mean \(\pm\) one standard deviation across the 10 evaluation seeds.}
  \label{fig:shift-3cities}
\end{figure*}

\section{Results}
\subsection{Trade-off frontier and runtime tunability}
We evaluate a single preference-conditioned policy across a sweep of $w_{\mathrm{bus}}$ and compare against fixed-time, a rule-based TSP overlay, and fixed-weight MaskablePPO specialists.
Figure~\ref{fig:pareto_tradeoff} shows multiple non-dominated MORL points across the three case-study intersections. The learned policy traces a smooth empirical trade-off frontier between $J_b$ and $J_{nb}$.
Table~\ref{tab:case_studies_wide} presents the quantitative data underlying the figure, through a frontier quality overview and operating-point outcomes at $w_{\mathrm{bus}}\in\{0.3,0.5,0.7\}$ across the 10 evaluated seeds.
Compared with PPO-S, MORL has slightly higher HV on all three case studies, with a favorable operating-point profile on LA2114 and NYC10802; on CHI2412, it achieves lower bus delays at the cost of higher non-bus and tail delays.
At moderate preferences ($w_{\mathrm{bus}}=0.3$--$0.7$), MORL reduces both bus and non-bus delay relative to the fixed-time and rule-based baselines across all three case studies.

\subsection{Distribution-shift robustness}\label{sec:shift}
Figure~\ref{fig:shift-3cities} compares in-distribution evaluation (see Section~\ref{sec:regimes}) to the default-IZ distribution shift for the same three intersections. The frontier shapes remain broadly similar in these case studies, while absolute delay levels shift.
The shift is most visible in bus-delay outcomes under lower-traffic, lower-bus-prevalence conditions, consistent with the controller having more feasible opportunities to accelerate buses when there are fewer conflicting priority requests.
This provides preliminary evidence that runtime tuning can transfer to lower bus-prevalence regimes, with scenario-dependent performance offsets.

\subsection{Operating points and tail-delay externalities}

\begin{figure}
  \centering
  \includegraphics[width=\columnwidth]{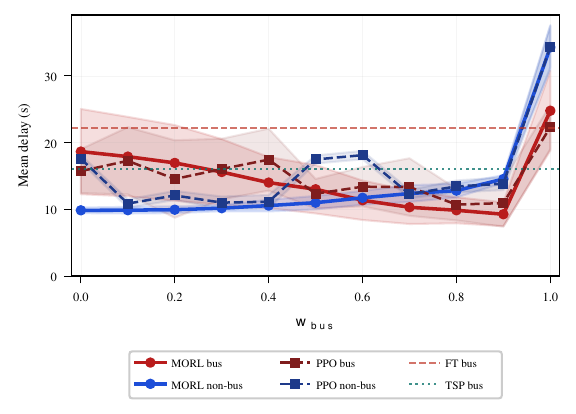}
  \caption{Runtime tunability on NYC10802: mean bus and non-bus delay as a function of $w_{\mathrm{bus}}$ for the conditioned policy and PPO fixed-weight agents, with fixed-time and rule-based TSP bus-delay reference levels. Shaded bands denote mean \(\pm\) one standard deviation across the 10 evaluation seeds.}
  \label{fig:tunability-NYC10802}
\end{figure}

On NYC10802, $w_{\mathrm{bus}}$ serves as a practical \emph{runtime operating knob}: increasing $w_{\mathrm{bus}}$ smoothly reduces bus delay while increasing non-bus delay (Figure~\ref{fig:tunability-NYC10802}). For example, moving from $w_{\mathrm{bus}}{=}0.3$ to $0.7$ lowers mean bus delay from $16.6$s to $11.3$s, while mean non-bus delay rises from $10.1$s to $12.3$s (Table~\ref{tab:case_studies_wide}). Tail impacts are comparatively mild in this case: $J^{\mathrm{tail}}_{\mathrm{nb}}{=}\mathrm{CVaR}_{0.1}$ increases only from $57.9$s to $61.5$s over the same range (Table~\ref{tab:case_studies_wide}). PPO specialists perform well at their trained setpoints but offer only discrete operating points and do not provide the same continuous tunability without training multiple policies.

\begin{table*}
\topfloatguard
\centering
\scriptsize
\setlength{\tabcolsep}{2.8pt}
\renewcommand{\arraystretch}{1.15}
\caption{Additional intersections (evaluation protocol in Section~\ref{sec:exp}). MORL Pareto-set quality is summarized by HV\% and \#ND over the 11-point preference sweep ($w_{\mathrm{bus}}\in\{0.0,0.1,\dots,1.0\}$). FT-dom and TSP-dom report the mean$\pm$std number of preferences (out of 11) where MORL dominates FixedTime or RuleTSP. Baseline delays are reported as $(J_b, J_{\mathrm{nb}})$ (mean$\pm$std). MORL ranges report the min/max of the mean delay across preferences.}
\label{tab:additional_10cities}
\resizebox{\textwidth}{!}{%
\begin{tabular}{lrrrrrrrr}
\toprule
Tag & \#ND & HV\% & FT-dom & TSP-dom & FixedTime $(J_b,J_{\mathrm{nb}})$ & RuleTSP $(J_b,J_{\mathrm{nb}})$ & MORL $J_b$ range & MORL $J_{\mathrm{nb}}$ range \\
\midrule
atl66 & 7.9$\pm$1.6 & 96.1$\pm$0.4 & 9.9$\pm$1.9 & 8.3$\pm$2.9 & (19.3$\pm$4.4, 21.5$\pm$0.8) & (15.8$\pm$3.7, 20.8$\pm$0.6) & [10.8--18.0] & [9.5--15.9] \\
bos666 & 2.5$\pm$0.8 & 83.3$\pm$1.1 & 8.9$\pm$1.9 & 9.6$\pm$0.7 & (114.8$\pm$88.9, 214.3$\pm$6.6) & (99.9$\pm$54.9, 202.8$\pm$8.2) & [32.6--71.7] & [60.5--245.9] \\
chi1027 & 3.5$\pm$2.1 & 82.7$\pm$1.1 & 10.8$\pm$0.4 & 10.7$\pm$0.5 & (141.1$\pm$12.3, 152.8$\pm$16.1) & (109.6$\pm$11.7, 135.8$\pm$20.8) & [43.2--57.3] & [54.0--106.0] \\
chi758 & 2.7$\pm$1.5 & 76.6$\pm$2.5 & 1.0$\pm$2.2 & 0.0$\pm$0.0 & (72.8$\pm$17.6, 70.3$\pm$13.3) & (59.2$\pm$13.5, 67.4$\pm$10.8) & [52.0--155.4] & [81.0--282.9] \\
dal1082 & 3.3$\pm$0.9 & 82.7$\pm$1.6 & 11.0$\pm$0.0 & 11.0$\pm$0.0 & (126.8$\pm$37.0, 132.1$\pm$2.9) & (114.0$\pm$27.3, 130.5$\pm$3.9) & [37.9--54.0] & [55.7--69.9] \\
la477 & 3.2$\pm$1.4 & 52.1$\pm$3.4 & 8.8$\pm$1.8 & 9.1$\pm$1.2 & (233.8$\pm$80.7, 233.0$\pm$20.2) & (241.6$\pm$82.7, 234.4$\pm$18.3) & [107.8--194.2] & [181.4--266.6] \\
nyc1078 & 3.1$\pm$1.3 & 82.0$\pm$2.5 & 9.6$\pm$2.4 & 9.3$\pm$2.7 & (164.6$\pm$89.3, 304.6$\pm$102.9) & (129.6$\pm$69.4, 288.2$\pm$93.7) & [32.2--126.9] & [64.8--283.7] \\
sea543 & 3.7$\pm$1.6 & 49.5$\pm$3.8 & 3.7$\pm$3.8 & 3.5$\pm$4.1 & (161.7$\pm$171.9, 290.1$\pm$8.7) & (160.4$\pm$189.7, 288.1$\pm$8.2) & [98.5--199.9] & [208.4--348.5] \\
sf838 & 3.2$\pm$1.3 & 86.6$\pm$0.8 & 10.7$\pm$0.9 & 10.1$\pm$2.8 & (84.8$\pm$47.4, 214.2$\pm$2.8) & (73.7$\pm$39.4, 201.5$\pm$11.1) & [31.7--40.4] & [39.0--48.9] \\
slc47 & 3.2$\pm$1.5 & 66.8$\pm$3.4 & 9.4$\pm$0.7 & 7.1$\pm$3.3 & (138.1$\pm$39.5, 146.6$\pm$4.1) & (113.2$\pm$36.9, 148.0$\pm$3.5) & [82.9--124.2] & [111.6--205.6] \\
\bottomrule
\end{tabular}}
\end{table*}


\sisetup{round-mode=places,round-precision=2}
\newcommand{\heat}[1]{%
  \pgfmathsetmacro{\absrho}{abs(#1)}%
\pgfmathsetmacro{\pct}{ifthenelse(\absrho<0.45,0,min(100,30*\absrho))}
  \ifdim #1 pt > 0pt
    \edef\heatcolor{blue!\pct!white}%
  \else
    \edef\heatcolor{red!\pct!white}%
  \fi
  \expandafter\cellcolor\expandafter{\heatcolor}\num{#1}%
}

\begin{table}
\centering
\scriptsize
\setlength{\tabcolsep}{2.5pt}
\renewcommand{\arraystretch}{1.15}
\caption{Spearman rank correlations ($\rho$) between MORL performance summaries and static intersection features across 13 intersections. Cell color encodes sign and magnitude (blue=positive, red=negative). Significance testing is omitted/handled separately due to multiple comparisons and small $n$.}
\label{tab:heat_map}
\begin{tabular}{lrrrr}
\toprule
Feature & HV\% & \# ND & FT Dom & TSP Dom \\
\midrule
Total Flow & \heat{-0.46} & \heat{-0.03} & \heat{0.03} & \heat{0.11} \\
Traffic Imb. & \heat{-0.33} & \heat{-0.41} & \heat{-0.42} & \heat{-0.25} \\
\#App. & \heat{-0.37} & \heat{-0.21} & \heat{-0.60} & \heat{-0.54} \\
\#Phases & \heat{-0.62} & \heat{-0.51} & \heat{-0.89} & \heat{-0.64} \\
\#Lanes & \heat{-0.28} & \heat{-0.07} & \heat{-0.52} & \heat{-0.37} \\
Cycle (s) & \heat{-0.60} & \heat{-0.54} & \heat{-0.74} & \heat{-0.51} \\
Green Imb. & \heat{-0.35} & \heat{-0.40} & \heat{-0.42} & \heat{-0.24} \\
Max Phase Share & \heat{0.26} & \heat{-0.02} & \heat{0.51} & \heat{0.46} \\
\bottomrule
\end{tabular}
\end{table}

\subsection{Across-intersection reproducibility}\label{sec:across}

To assess whether the method exhibits consistent behavior beyond the three case-study intersections, we evaluate the same method on 10 additional intersections spanning multiple cities and intersection geometries. We train a separate conditioned policy per intersection using a fixed set of hyperparameters and preference curriculum, with no per-site hyperparameter retuning. Table~\ref{tab:additional_10cities} reports frontier summaries, baseline comparisons, and MORL delay ranges across these sites.

The results show substantial variation in frontier quality, with HV\% ranging from 49.5 (SEA543) to 96.1 (ATL66) and \#ND ranging from about 2.5 to 7.9. MORL dominates FixedTime and RuleTSP for most preferences in many intersections (FT-dom/TSP-dom near 9--11), but performance is weaker in some cases such as CHI758 (FT-dom 1.0, TSP-dom 0.0), where MORL ranges are wide and trade-offs are steep. Overall, the additional-intersection results show consistent, preference-dependent trade-offs under route/timetable resampling, but also indicate that frontier quality is intersection-dependent.

\subsection{Structural correlates of frontier quality}
Table~\ref{tab:heat_map} suggests that frontier quality and dominance decrease with intersection complexity. HV\%, \#ND, and dominance generally correlate negatively with the number of approaches, phases, lanes, and cycle length, with the strongest negative correlations for phase count and dominance (down to \(\rho\approx -0.89\)). The max phase share correlates positively with dominance (up to \(\rho\approx 0.51\)). This analysis is exploratory and hypothesis-generating only; due to the small sample size and multiple comparisons, no causal interpretation or claim of statistical significance is made.

\section{Discussion and Limitations}

Results show that a single conditioned policy can cover a broad trade-off region and often dominate fixed-time and rule-based baselines at moderate preferences, but tunability is not uniform: non-bus delay and tail-delay can increase sharply at high bus weights, and some intersections exhibit low HV and few non-dominated points, indicating harder control regimes. Operationally, $w_{\mathrm{bus}}$ should be treated as a bounded knob: moderate settings capture most bus-delay gains with limited externalities, while high-priority settings can yield diminishing returns and increase non-bus and tail delays; high $w_{\mathrm{bus}}$ should therefore be reserved for exceptions and paired with guardrails (e.g., monitoring $J_{nb}$ and CVaR tail-delay and rolling back if thresholds are exceeded). Fixed-weight specialists can still be stronger at exact setpoints, reflecting a specialization--generalization trade-off.

Limitations include dependence on scenario realism (demand, bus prevalence, timetable variability), restriction to single-intersection control, and the use of linear scalarization, which may miss non-convex parts of the Pareto set. Distribution shift evaluation is limited to bus-prevalence changes, and the structural correlation analysis is descriptive given small $n$ and uncorrected multiple comparisons.

\section{Conclusion}
We presented a runtime-tunable TSP framework based on preference-conditioned MORL, enabling operators to adjust the transit-vs-traffic trade-off at evaluation time with a single trained controller. By extending IZ with a constrained phase-control/TSP wrapper, bus-prevalence augmentation and timetable insertion wrappers, and a rule-based TSP overlay baseline, we provide a standardized and operationally grounded evaluation setting for RL-based TSP. Across 13 intersections, the conditioned-policy approach (one policy trained per intersection) traces a trade-off frontier, often dominates fixed-time and rule-based baselines at moderate preferences, and exhibits qualitative robustness under a bus-prevalence shift, while tail-delay diagnostics expose the externalities of high-priority settings. Future work should extend to network-level control, richer preference models, and more diverse distribution shifts to further validate deployability.

\bibliographystyle{IEEEtran}
\bibliography{refs}

\end{document}